\documentclass[letterpaper, 10pt, conference]{ieeeconf}  
\usepackage{mathtools}
\usepackage{color}
\usepackage{array}
\usepackage{stfloats}
\usepackage[normalem]{ulem}
\usepackage{amssymb}
\usepackage{float}
\usepackage{nccmath}
\usepackage{graphicx}
\usepackage[linesnumbered,ruled]{algorithm2e}
\usepackage{setspace}
\usepackage{booktabs}
\usepackage{subcaption}
\usepackage{mwe}
\usepackage{cite}
\usepackage{amssymb,amsfonts}
\usepackage{graphicx}
\usepackage{textcomp}
\usepackage{xcolor}

\usepackage{algorithmicx}
\usepackage{xspace}
\usepackage{balance}
\usepackage{graphicx}  
\usepackage{subcaption}
\usepackage{kotex}
 
\newcommand{\argmax}{\mathop{\mathrm{argmax}}\limits} 

\def\BibTeX{{\rm B\kern-.05em{\sc i\kern-.025em b}\kern-.08em
    T\kern-.1667em\lower.7ex\hbox{E}\kern-.125emX}}

\IEEEoverridecommandlockouts                              

\title{\LARGE \bf
Multi-Agent Deep Reinforcement Learning using Attentive Graph Neural Architectures for Real-Time Strategy Games}

\author{Won Joon Yun$^{1}$, Sungwon Yi$^{2}$, and Joongheon Kim$^{1}$
\thanks{$^{1}$Won Joon Yun and Joongheon Kim are with the School of Electrical Engineering, Korea University, Seoul, Republic of Korea
        {\tt\small ywjoon95@korea.ac.kr, joongheon@korea.ac.kr}}%
\thanks{$^{2}$Sungwon Yi is with Electronics and Telecommunications Research Institute (ETRI), Daejeon, Republic of Korea
        {\tt\small sungyi@etri.re.kr}}%
}

\begin{document}

\maketitle
\thispagestyle{empty}
\pagestyle{empty}

\begin{abstract}
In real-time strategy (RTS) game artificial intelligence research, various multi-agent deep reinforcement learning (MADRL) algorithms are widely and actively used nowadays. Most of the research is based on \textit{StarCraft II} environment because it is the most well-known RTS games in world-wide.
In our proposed MADRL-based algorithm, distributed MADRL is fundamentally used that is called QMIX. In addition to QMIX-based distributed computation, we consider state categorization which can reduce computational complexity significantly. Furthermore, self-attention mechanisms are used for identifying the relationship among agents in the form of graphs. Based on these approaches, we propose a categorized state graph attention policy (CSGA-policy). As observed in the performance evaluation of our proposed CSGA-policy with the most well-known \textit{StarCraft II} simulation environment, our proposed algorithm works well in various settings, as expected. 
\end{abstract}


\section{Introduction}\label{sec:1}
In modern artificial intelligence (AI) and deep learning (DL) research~\cite{tg19liu,arxivpark,ijcnn19kim,icdcs18ahn}, there are a lot of research results for developing autonomous, intelligent, and cooperative real-time strategy (RTS) game players. Among various attractive game AI research and development trends, one of major research contributions is for the design and implementation of computer game players for \textit{StarCraft II}, which is the most successful world-wide RTS games~\cite{sc1}.

In order to build AI/DL-based RTS game players, reinforcement learning (RL) and deep reinforcement learning (DRL)-based algorithms are widely used because RL/DRL-based algorithms are fundamentally for stochastic and sequential decision making in order to achieve their own goals under uncertainty observations. Therefore, the RL/DRL-based algorithms are also actively used for wireless and mobile networks~\cite{iotj20kwon,jcn20kwon,twc19choi,IEEESMC20_MADRL1}, autonomous driving~\cite{ijcnn19shin,ijcai19shin,IEEESMC_MADRL2}, robotics~\cite{pomdp,kober2012reinforcement}, 
and so forth.
Furthermore, in order to consider the RTS game environment which is with multiple game players, multi-agent DRL (MADRL) should be considered which is much more complicated than single-agent DRL~\cite{globecom19kwon,busoniu2008comprehensive}.


There exist a lot of research contributions in MADRL, and the corresponding algorithms can be categorized into three parts, i.e., centralized MADRL, fully decentralized MADRL and centralized training and decentralized execution (CTDE) MADRL~\cite{horgan2018distributed,busoniu2008comprehensive,henderson2017deep,xuan2002multi,es,globecom19kwon}.

In centralized-MADRL, the state information of all agents is collected in a centralized server such as cloud. Since centralized-MADRL is trained with complete state information, the performance of policy is guaranteed. However,  centralized-MADRL has several challenges in real-life application. First, the relationship between the number of agents and the joint action space requires exponential complexity computation during training and execution. In addition, malfunctioned agents exist in various lives, however the computation time does not decrease. According to the aforementioned overheads, the concept of fully decentralized MADRL is re-examined.

Besides the centralized MADRL-based AI/DL algorithms, there is fully decentralized MADRL which is the opposite concept of centralized MADRL. The fully decentralized MADRL operates well without concerning the centralized MADRL problem (e.g., the exponential complexity of joint action space, or the situation of malfunctioned agents).
The origin of decentralized MADRL-based algorithms is the independent Q-learning (IQL)~\cite{icml93tan,mnih2013playing} which is fundamentally based on partially observable Markov decision process (POMDP)~\cite{pomdp}. To describe the mechanism in nutshell, every single agent in IQL has its own policy and it makes decision for its own action in same environment. 
Because each agent has its own unique reward function, the agents take greedy actions in order to maximize their own reward without the consideration of the other agents' rewards. Therefore, it is obvious that IQL-based approach is not good for multi-agent cooperation and coordination. 

In summary, there exists a dilemma between centralized MADRL and fully decentralized MADRL model. The CTDE method is proposed in order to deal with the dilemma by combining the advantage of two methods (i.e., centralized and decentralized). The CTDE method has two different procedures; 1) centralized training and 2) decentralized execution. Every agent has its own policy and make decisions with its own policy, which is similar to IQL. However, the observations and policies of all agents are trained with the centralized method for cooperation or coordination of agents. There are two main algorithms of CTDE MADRL which are based on a value decomposition network (VDN) for cooperative behaviors among multiple agents~\cite{aamas18sunehag}. The VDN-based algorithm shows that the joint action-value function can be factorized with the summation of the all action-value functions in all agents.
The proposed algorithm in~\cite{QMIX} assumes that the joint action-value function can be factorized with monotonic mixing network (QMIX). 
These two algorithms, i.e., VDN and QMIX, are able to achieve successful factorization, moreover, they outperforms IQL-based algorithms as well-studied in~\cite{zhou2019factorized}. 

When the decentralized or CTDE MADRL-based policy takes state information as an input, the state is generally composed of not only the characteristics of agents but also the partial states of other agents or other objects. 
If the policy is designed with deep neural network architecture based multi-layer perceptron (i.e., dense layers)~\cite{icufn17}, the units within the layers obtain every element of the given state. However, this learning mechanism is not efficient because of the unnecessary combination of state information. For that reason, this paper proposes a novel method which categorizes states. For our specific \textit{StarCraft II} RTS game applications, the state categorization can be performed and three categories can be generated, i.e., the characteristics of agents, partial information of other agents, and partial information of 
enemies. In addition, each state is encoded to be inputs for categorized layers. The encoded states are mapped into the agents which comprise graph models, then the attention-based graph neural architectures identify the relationship among agents~\cite{scarselli2008graph,zhang2019heterogeneous}.

The main contributions of this research can be summarized as following three aspects.
\begin{itemize}
    \item First of all, a new method for categorized states is introduced based on the fact that input state information can be categorized. 
    \item In addition, the categorized state can be mapped into agents, and then edges in graph representation can be identified via self-attention. Finally, the distributed MADRL actions can be derived by the graph representation. 
    \item Lastly, our proposed distributed MADRL-based RTS game algorithm in this paper is evaluated in the most well-known \textit{StarCraft II} RTS game environment that is called SMAC where the SMAC stands for StarCraft Multi-Agent Challenge.  
\end{itemize}

The rest of this paper is organized as follows.
First, we introduce the background of decentralized-MADRL in Sec.~\ref{sec:2} in order to describe our proposed architecture. 
Then, our proposed categorized state graph attention policy (CSGA-policy) algorithm is described in Sec.~\ref{sec:3}.
In addition, the proposed architecture in \textit{StarCraft II} is evaluated that is real-time strategy game applications in Sec.~\ref{sec:4}. 
Lastly, Sec.~\ref{sec:5} concludes this paper.


\section{Related Work}\label{sec:2}

This section explains major two related neural architectures, i.e., attentive graph neural network (refer to Sec.~\ref{sec:agnn}) and cooperative MADRL (refer to Sec.~\ref{sec:cMADRL}).

\subsection{Attentive Graph Neural Architectures}\label{sec:agnn}
\subsubsection{Attention Mechanism}
An attention mechanism is widely used in various fields such as natural language processing, computer vision, distributed learning, an so forth~\cite{attention}.
With this attention mechanism, suppose that any types of inputs are given. Then, the input can be transformed as a context vector through encoding process. This context vector helps outputs perform well by assigning dynamic weights.
Among various attention mechanisms, a scale-dot attention (i.e., query, key, and value attention) is one of well-known self-attention mechanisms. 
If the input $X\triangleq\left\{x_1,x_2,\cdots,x_N\right\}$ passes through the query layer $q(\cdot)$, key layer $k(\cdot)$ and value layer $v(\cdot)$, the outputs are represented as follows.
\begin{eqnarray}
q(X) &\triangleq& \left\{q(x_1),q(x_2),\cdots,q(x_N)\right\},\label{eq:soft1} \\ 
k(X) &\triangleq& \left\{k(x_1),k(x_2),\cdots,k(x_N)\right\},\label{eq:soft2} \\
v(X) &\triangleq& \left\{v(x_1),v(x_2),\cdots,v(x_N)\right\}\label{eq:soft3}.
\end{eqnarray}

For $\forall n\in[1,N]$ and $\forall m\in[0,N]\backslash n$, the similarity between input $x_n$ and $x_m$ can be obtained as follows,
\begin{equation}
    a_{n,m} = \frac{q(x_n)\cdot k(x_m)}{\sqrt{N}}*v(x_m)\label{eq:soft4}{\color{blue}{.}} 
\end{equation} 
where $a_{n,m}$ stands for attention scores.

A hard attention is a kind of stochastic attention mechanisms that is usually utilized in RL/DRL formulation. The hard attention is simpler than self-attention mechanisms because this hard attention returns only binary vector which shows whether the output is correct or not.
\subsubsection{Attentive Graph Neural Architectures} 
A graph can be represented as $G({V},{E})$ where $V$ and $E$ stand for vertices and edges, respectively. If ${V}$ is given and the relation among ${V}$ is not defined, a scale-dot attention is able to easily reveal the edges ${E}$ with the form of an adjacent matrix. In addition, with the hard attention of vertices ${V}$, the binary adjacent matrix can be obtained where the binary adjacent matrix can be implemented as a mask. Based on the scale-dot attention and hard attention mechanisms, $G({V},{E})$ can be finally defined.

\subsection{Cooperative MADRL}\label{sec:cMADRL}
\begin{figure*}[t!]
\centering
\setlength{\tabcolsep}{2pt}
\renewcommand{\arraystretch}{0.2}
\includegraphics[page=1, width=0.99\linewidth]{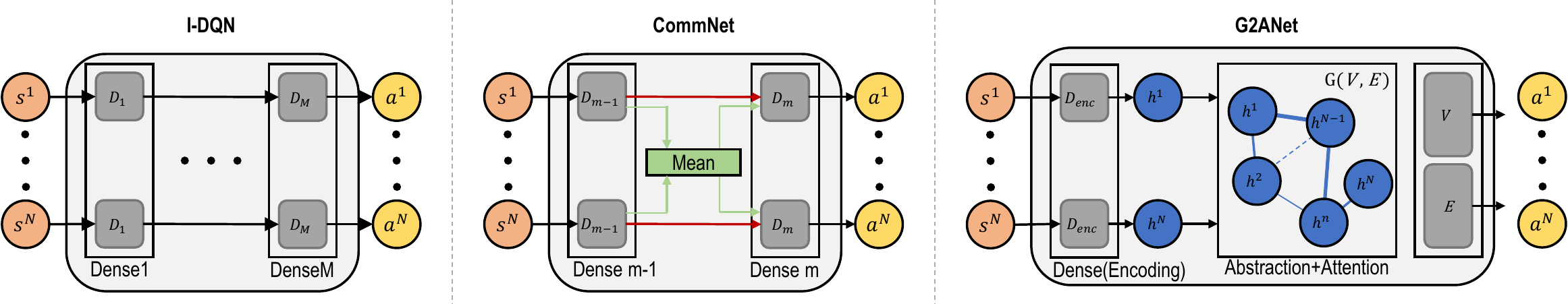}
\caption{Comparison among various MADRL policies.}
\label{fig:1}
\end{figure*}
\begin{figure*}[t!]
\centering
\setlength{\tabcolsep}{2pt}
\renewcommand{\arraystretch}{0.2}
   \includegraphics[page=1, width=0.9\linewidth]{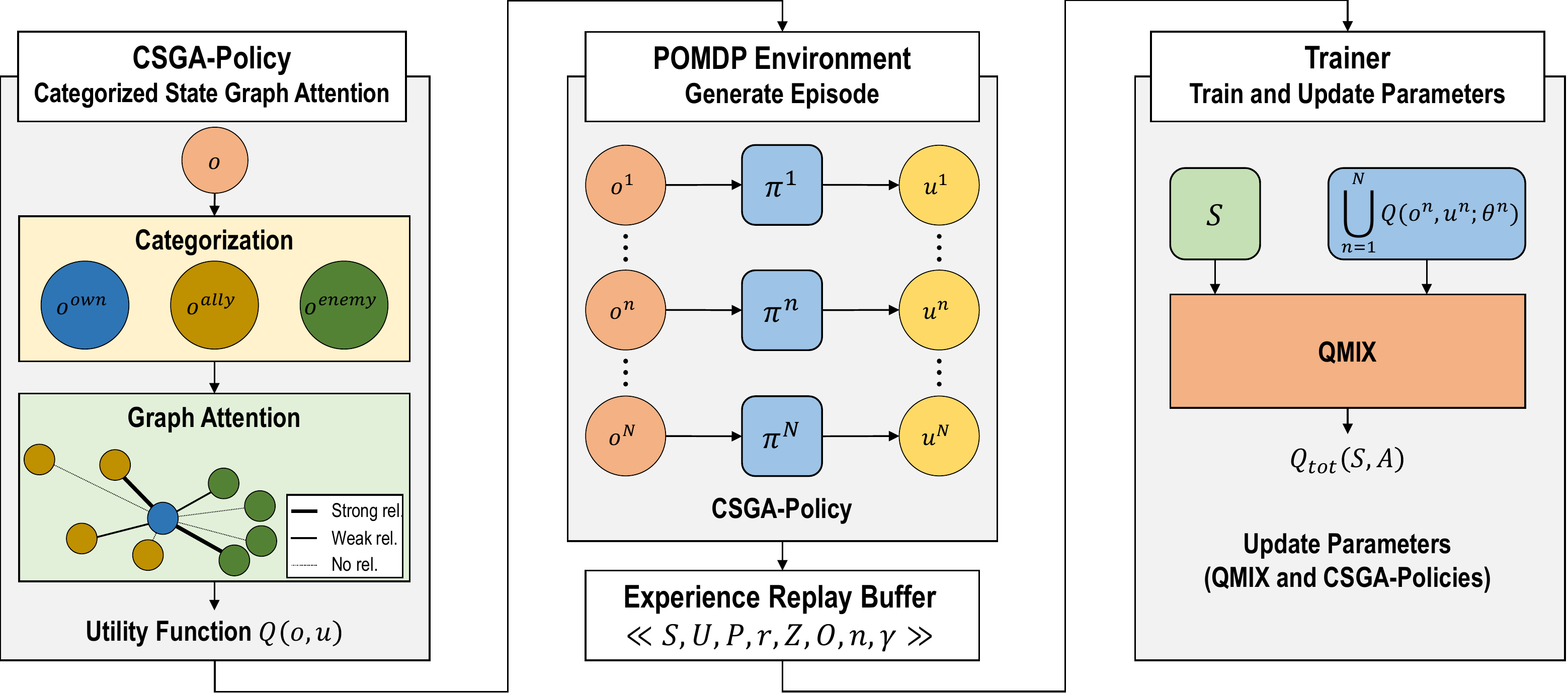}
\caption{Proposed MADRL-based algorithm design including CSGA-policy and QMIX neural architectures.}
\label{fig:2}
\end{figure*}

In this section, we introduce the preliminaries of the proposed architecture in terms of multi-agents' cooperation. In independent deep Q-network (I-DQN)~\cite{foerster2016learning}, each agent $n$ is able to obtain its own observation $o^n_t$. Based on the observation, each agent decides its own action $u^n_t$ via its own policy $\pi(o^n_t, u^n_t;\theta^n)$ in same environment. However, the policies of agents are all independent among them, thus the behaviors of agents are independent as well~\cite{icml93tan}.

In centralized-MADRL, on the other hand, there are some policies for cooperative actions. In particular, CommNet policy makes agents be cooperative by sharing communication layers among all agents~\cite{icoin21jung,tii20shin}. Furthermore, G2ANet policy decides cooperative actions through the abstraction to graph and attention mechanism in order to represent the relationship among agents~\cite{g2anet}.

Fig.~\ref{fig:1} shows the difference of I-DQN~\cite{icml93tan,mnih2013playing}, CommNet~\cite{icoin21jung,tii20shin} and G2ANet~\cite{g2anet}. All inputs $S=\{s^1,\cdots,s^N\}$ in Fig.~\ref{fig:1} are the states those are independent. In I-DQN, it takes an independent operation for each state. Because the dense layer takes only one vector for individual computation, the actions in each agent do not consider the actions of the others. 
On the other hand, a CommNet-based MADRL neural architecture~\cite{icoin21jung,tii20shin} has the communication layer that takes the \textsf{mean} operation of the other agents' hidden variables. For this reason, the average value of the other agents' hidden variables can affect on our considering agent's action. Therefore, we can observe that the actions are cooperative. 
In this CommNet-based all agents are homogeneous, thus sophisticated relationship among agents cannot be clearly represented.
In case of G2ANet-based MADRL neural architectures, the agents' hidden variables are mapped into vertices $V$ via an encoding layer that is denoted as $f^{\texttt{enc}}$, then the vertices get attention block which can be used for calculating the weighted edges $E$. Then, $G(V,E)$ can be obtained from the attention block. Based on this graph-based structure, more complicated and sophisticated relationships among agents can be numerically identified comparing to I-DQN and CommNet based MADRL neural architectures.

In addition to CommNet-based and G2ANet-based MADRL algorithms, there exists a QMIX-based MADRL architecture which is known as semi-distributed MADRL~\cite{QMIX}. 
In this QMIX-based MADRL architecture, there exists a common reward, that is defined as joint-action value function. Here, the joint action-value function is as follows,
\begin{equation}
    Q_{\textsf{total}}\left(s,u^1,\cdots,u^N;\theta\right) = \bigcup^{n=N}_{n=1}\left\{\argmax_{u^n} Q^n(o^n,u^n)\right\}.
    \label{eq:jav}
\end{equation} 

During the training process, the learner selects actions $U=\{u^1,\cdots,u^N\}$ which can maximize the joint action-value function~\eqref{eq:jav} and its own utility function $Q^n(o^n,u^n)$, simultaneously. For the $\argmax$ computation in \eqref{eq:jav}, the QMIX MADRL neural architecture should satisfy following condition,
\begin{equation}
    \frac{\partial Q_{\textsf{total}}}{\partial Q^{n}} \geq 0, \quad\forall n \in [1, N].
\end{equation}

Aforementioned process only proceeds in training policies in a centralized manner. Aer this training phase, each QMIX-based MADRL agent can inference its own policy in a distributed manner, i.e., each agent $n$ transfers its own observation $o^n$ to its fine-tuned policy $\pi^n$ and returns cooperative actions $u^n$.
Note that the QMIX-based MADRL algorithms can be called semi-distributed because it is centralized during training whereas it is distributed during inference. 

\section{Categorized State Graph Attention Policy (CSGA-Policy)}\label{sec:3}
In general MADRL formulations, each agent's state observation consists of the agent's unique state, other agents' partial observation, and the general state information of environment. 
According to the fact that \textit{StarCraft II} is an RTS game, the MADRL-based game agent's state information can be categorized into agent's own observation, allies (the other agents)' partial information, and enemies' information~\cite{DBLP:journals/corr/abs-1708-04782}. After the separation of states into aforementioned several categories, graph attention methods can be utilized for defining and learning the correlation and relationship. 
Our proposed MADRL-based algorithm in this paper is fundamentally based on the state categorization and attention-based graphs, thus, the algorithm is named to a \textit{Categorized State Graph Attention Policy (CSGA-Policy)}.

\begin{figure*}[t!]
\centering
\setlength{\tabcolsep}{2pt}
\renewcommand{\arraystretch}{0.2}
\includegraphics[page=1, width=0.99\linewidth]{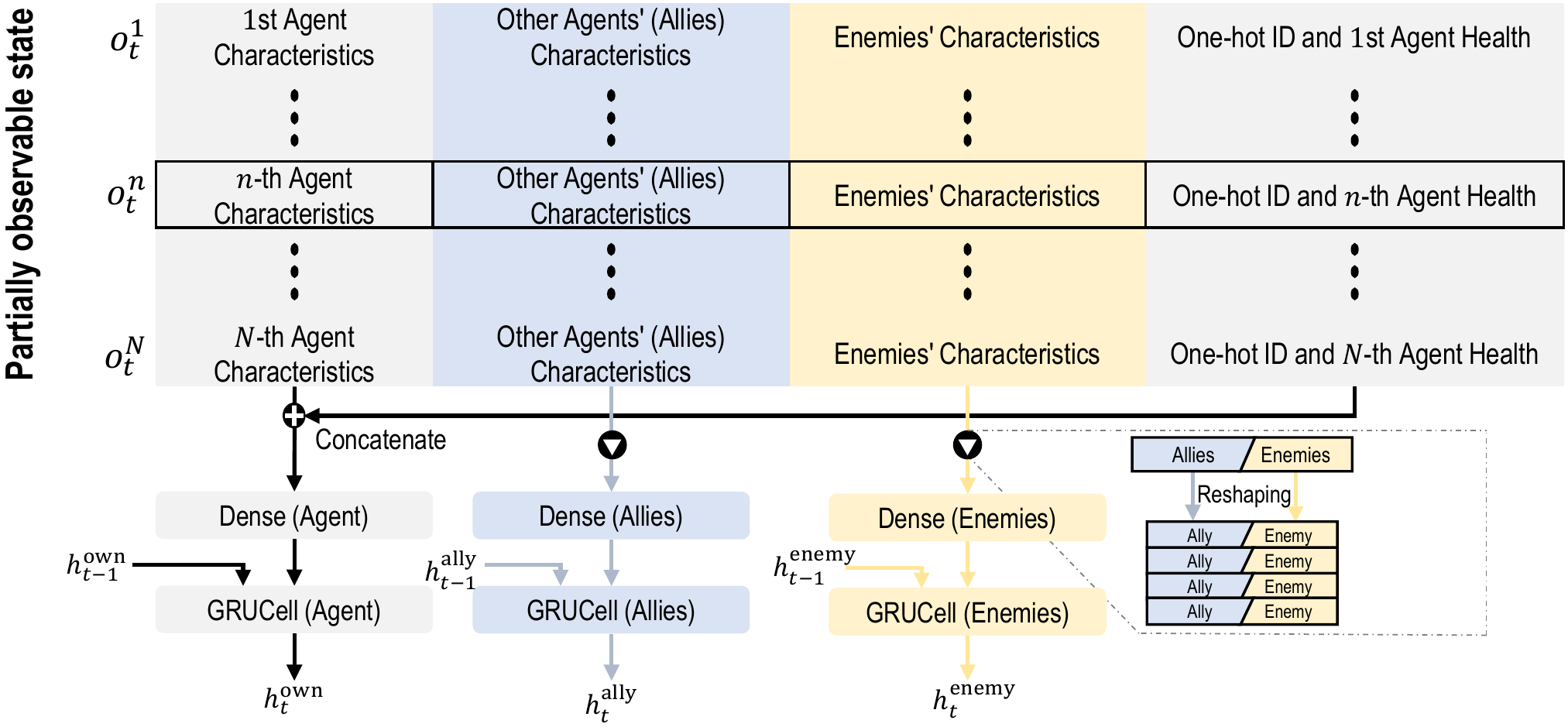}
\caption{Encoding and categorization processes (Step 1).}
\label{fig:3}
\end{figure*}

\subsection{System Model}
Our proposed MADRL neural architecture CSGA-policy is fundamentally based on deep recurrent Q-network (DRQN)~\cite{hausknecht2015deep}. To utilize the hard attention and scale-dot attention with graphs, state categorization is used. Each MADRL agent has its own CSGA-policy. In the learning phase in each agent, individual agents are trained for cooperative actions using the QMIX neural architecture. Note that the QMIX-based training is centralized whereas the inference in distributed in each MADRL agent. Fig.~\ref{fig:3} shows the corresponding learning system model.
\subsubsection{Categorization} SMAC, i.e., the \textit{StarCraft II} library which provides $23$ different and various combat scenarios for MADRL research~\cite{samvelyan2019starcraft}. There are $N$-agents and $M$-enemies in the SMAC environment. The agents can observe its own features (i.e., id, movement, health, and shield, those are denoted as $o_{n}^{\textsf{id}}$, $o_{n}^{\textsf{move}}$, $o_{n}^{\textsf{health}}$, and $o_{n}^{\textsf{shield}}$, respectively, as shown in \eqref{eq:o2}). In addition, the agents can partially observe the features of other agents/allies (i.e., visibility, distance between agent and ally, relative $x$-position, relative $y$-position, health, and shield, those are denoted as ${a}^{\textsf{vis}}_{m}$, ${a}_{m}^{\textsf{dis}}$, ${a}_{m}^{x}$, ${a}_{m}^{y}$, ${a}_{m}^{\textsf{health}}$, and ${a}_{m}^{\textsf{shield}}$, respectively, as shown in \eqref{eq:o3}) and the features of enemies (i.e., visibility, distance between agent and ally, relative $x$-position, relative $y$-position, health, and shield, those are denoted as ${e}^{\textsf{vis}}_{m}$, ${e}_{m}^{\textsf{dis}}$, ${e}_{m}^{x}$, ${e}_{m}^{y}$, ${e}_{m}^{\textsf{health}}$, and ${e}_{m}^{\textsf{shield}}$, respectively, as shown in \eqref{eq:o4}). 

Each agent $n$ with the observation $o^n$ in \eqref{eq:o1} can be represented as~\eqref{eq:o1}--\eqref{eq:o4},
\begin{eqnarray}
\hspace{-6mm}    o^n &\triangleq& \left\{o^{n,\textsf{own}}, o^{n,\textsf{ally}},o^{n,\textsf{enemy}}\right\},
    \tag{O.1}\label{eq:o1}\\
\hspace{-6mm}    o^{n,\textsf{own}} &\triangleq& \left\{o_{n}^{\textsf{id}}, o_{n}^{\textsf{move}}, o_{n}^{\textsf{health}}, o_{n}^{\textsf{shield}}\right\}, \tag{O.2}\label{eq:o2}\\
\hspace{-6mm}    o^{n,\textsf{ally}} &\triangleq& \bigcup^{m=N}_{m\neq n,m \geq 1}\left\{{a}^{\textsf{vis}}_{m},{a}_{m}^{\textsf{dis}},{a}_{m}^{x},{a}_{m}^{y},{a}_{m}^{\textsf{health}},{a}_{m}^{\textsf{shield}}\right\}, \tag{O.3}\label{eq:o3}\\
\hspace{-6mm}    o^{n,\textsf{enemy}} &\triangleq& \bigcup^{l=N}_{l\neq n,l \geq 1}\left\{{e}^{\textsf{vis}}_{l},{e}_{l}^{\textsf{dis}},{e}_{l}^{x},{e}_{l}^{y},{e}_{l}^{\textsf{health}},{e}_{l}^{\textsf{shield}}\right\}, \tag{O.4}\label{eq:o4}
\end{eqnarray}
where $a^{\texttt{feature}}_{m}$ and $e^{\texttt{feature}}_{l}$ stand for \texttt{feature} of $m$-th ally and $l$-th enemy. 
The $o^n$ can be defined as the combination of following three, i.e., $o^{n,\textsf{own}}$, $o^{n,\textsf{ally}}$, and $o^{n,\textsf{enemy}}$, as shown in \eqref{eq:o1}.
Fig.~\ref{fig:3} illustrates the information that agents observe. 
As illustrated in Fig.~\ref{fig:3}, when $N-1$ allies and $L$ enemies exist, $o^{n,\textsf{ally}}$ and $o^{n,\textsf{enemy}}$ are transformed to ally matrix $\mathcal{A}^n$ (size: $(N-1)\times\texttt{dim(feature)}$) and enemy matrix $\mathcal{E}^n$ (size: $L\times\texttt{dim(feature)}$) , respectively, where $\texttt{dim(\textrm{\textit{k}})}$ is defined as a function to obtain the dimension of given vector $k$. 
At time $t$, $\mathcal{A}^{\textsf{enc},n}_{t}$ and $\mathcal{E}^{\textsf{enc},n}_{t}$ are defined as the results of encoding layer when the inputs are $\mathcal{A}^n$ and $\mathcal{E}^n$, respectively. 
The $\mathcal{A}^{\textsf{enc},n}_{t}$ and hidden layer parameter $h^{\textsf{ally}}_{t-1}$ will be the input of \texttt{ally-RNN}, and similarly, $\mathcal{E}^{\textsf{enc},n}_{t}$ and hidden layer $h^{\textsf{enemy}}_{t-1}$ will be the input of \texttt{enemy-RNN}, as well.
\subsubsection{Graph Attention}
There are two major methods for graph attentions, i.e., hard attention and scale-dot attention where the hard attention is for representing the relationship among nodes with $0$ or $1$ whereas the scale-dot attention is for determining the weights over the graph edges. 
The encoding process in Fig.~\ref{fig:3} takes three different types of input components, i.e., agent's its own hidden variable $h^{\textsf{own}}_t$, allies' their own hidden variable $h^{\textsf{ally}}_t$, and enemies' hidden variable $h^{\textsf{enemy}}_t$. 

\begin{figure*}[t!]
\centering
\setlength{\tabcolsep}{2pt}
\renewcommand{\arraystretch}{0.2}
\includegraphics[page=1, width=0.99\linewidth]{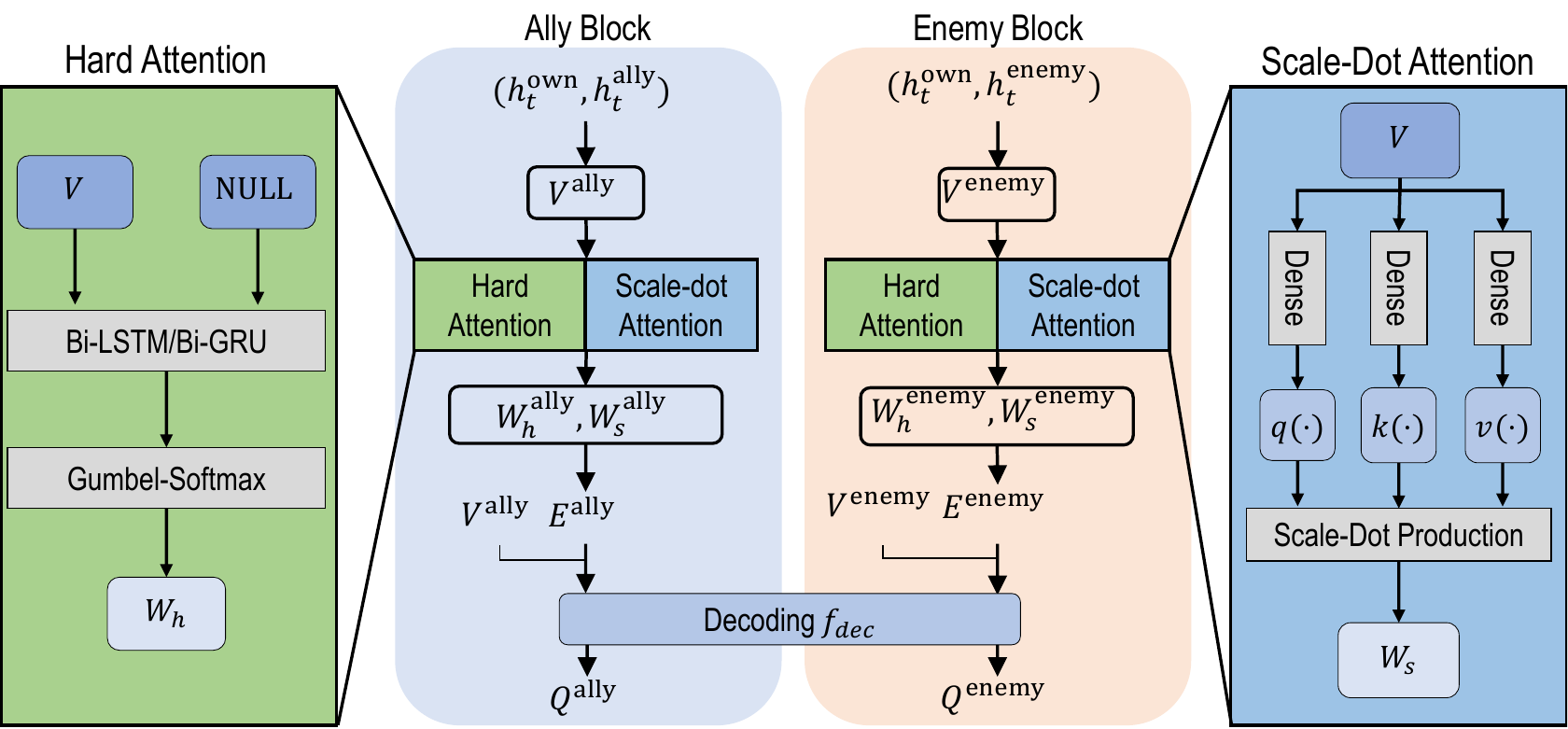}
\caption{Attention and decoding processes (Step 2).}
\label{fig:4}
\end{figure*}

Fig.~\ref{fig:4} illustrates the MADRL learning computational procedures for allies and enemies. 
The proposed MADRL-based algorithm gathers the information after state categorization, and separates the results into allies' information and enermies' information. After that, the results will be the inputs for Ally Block and Enemy Block in Fig.~\ref{fig:4}. 

In order to obtain \texttt{own-allies graph} and \texttt{own-enemies graph}, the propose method obtains $V^{\textsf{ally}}$ and $V^{\textsf{enemy}}$ at first, where the $V^{\textsf{ally}}$ and $V^{\textsf{enemy}}$ are defined as 
the $(N-1)\times$\texttt{dim(RNN)} matrix which is $N-1$ times replication of $h^{\textsf{own}}_t$ and 
the $L\times$\texttt{dim(RNN)} matrix which is $L$ times replication of $h^{\textsf{enemy}}_t$, respectively.
Then, these $V^{\textsf{ally}}$ and $V^{\textsf{enemy}}$ can be mapped to $N-1$ vertices and $L$ vertices, respectively.
Suppose that $\textsc{NULL}^{\textsf{ally}}$ and $\textsc{NULL}^{\textsf{enemy}}$ are the matrices where all elements are zero with the sizes of $(N-1)\times$\texttt{dim(RNN)} and $L\times$\texttt{dim(RNN)}. 

Based on these obtained
$V^{\textsf{ally}}$, $V^{\textsf{enemy}}$, $\textsc{NULL}^{\textsf{ally}}$, and $\textsc{NULL}^{\textsf{enemy}}$, following computation can be done in order to get the weight vectors $W_{h}^{\textsf{ally}}$ (size: $N-1$) and $W_{h}^{\textsf{enemy}}$ (size: $L$) where the values of vectors are $0$ and $1$,
\begin{eqnarray}
W_{h}^{\textsf{ally}}&=&\texttt{G-softmax}(\texttt{BiGRU}(V^{\textsf{ally}},\textsc{NULL}^{\textsf{ally}}))\\
W_{h}^{\textsf{enemy}}&=&\texttt{G-softmax}(\texttt{BiGRU}(V^{\textsf{enemy}},\textsc{NULL}^{\textsf{enemy}}))
\end{eqnarray}
where \texttt{G-softmax}($\cdot$) and \texttt{BiGRU}($\cdot$) stand for gumbel-softmax~\cite{gumbelsoftmax} and bidirectional GRU cell~\cite{tii19deng}, respectively. 
After this computation, the attention scores $Att^{\textsf{ally}}$ and $Att^{\textsf{enemy}}$ should be obtained via $V^{\textsf{ally}}$ (for \texttt{own-ally}) and $V^{\textsf{enemy}}$ (for \texttt{own-enemy})~~\eqref{eq:soft1}--\eqref{eq:soft4}, respectively. Finally, soft weights $W_s^{\textsf{ally}}$ and $W_s^{\textsf{enemy}}$ can be obtained applying \texttt{Softmax} function to the attention scores $Att^{\textsf{ally}}$ and $Att^{\textsf{enemy}}$.
Now, the weighted adjacency matrices $E^{\textsf{ally}}$, $E^{\textsf{enemy}}$ can be computed using element-wise product for the given $(W_h^{\textsf{ally}}$, $W_s^{\textsf{ally}})$ and $(W_h^{\textsf{enemy}}$, $W_s^{\textsf{enemy}})$.

\begin{figure*}[t!]
\centering
\setlength{\tabcolsep}{2pt}
\renewcommand{\arraystretch}{0.2}
\includegraphics[page=1, width=0.8\linewidth]{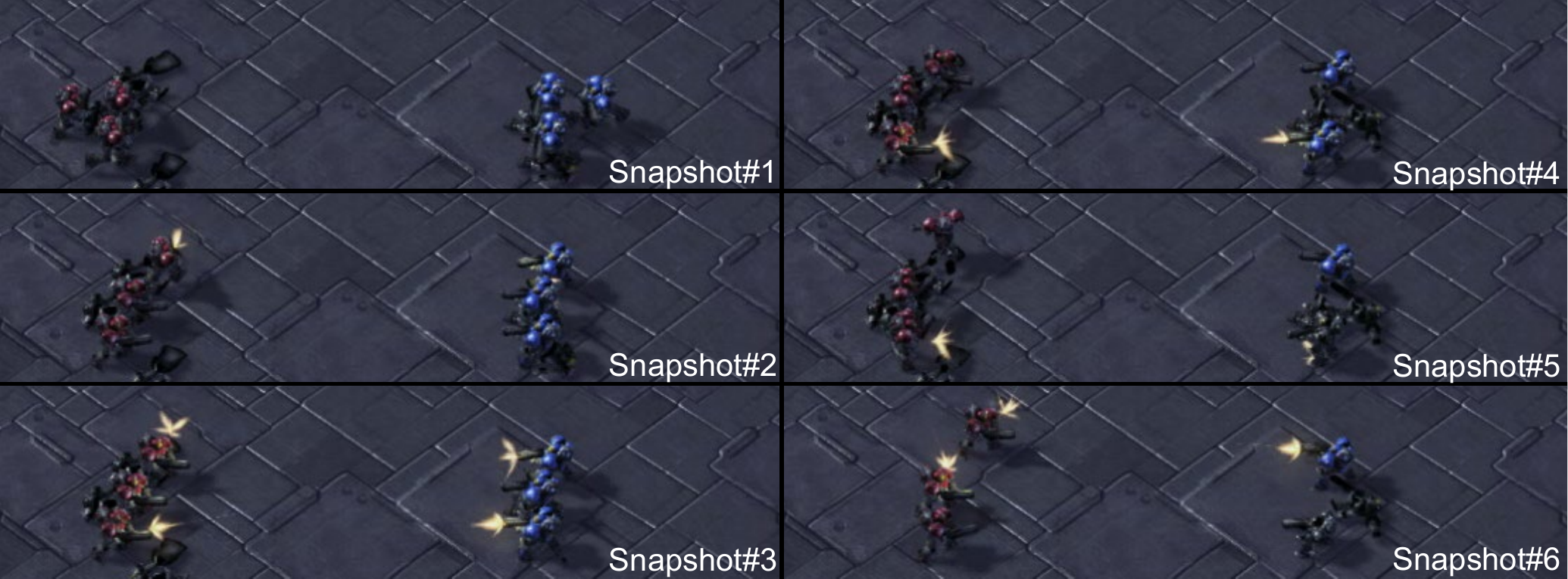}
\caption{MADRL performance evaluation results with \textsf{Simulation-3m} (3 marines vs. 3 marines).}
\label{fig:5}
\end{figure*}

Lastly, based on these results, the utility function as the results of decoding layers where the inputs are $G(V^{\textsf{ally}},E^{\textsf{ally}})$ and $G(V^{\textsf{enemy}},E^{\textsf{enemy}})$ is as follows,


\begin{eqnarray}
    Q^{\textsf{ally}}(o^{n},u^{n})&=&\frac{1}{N-1}f^{\texttt{dec}}(G(V^{\textsf{ally}},E^{\textsf{ally}})) \\
    Q^{\textsf{enemy}}(o^{n},u^{n})&=&\frac{1}{L}f^{\texttt{dec}}(G(V^{\textsf{enemy}},E^{\textsf{enemy}}))
\end{eqnarray}
where $f^{\texttt{dec}}$ stands for the functional representation of decoding layers and $u^n$ represents the action of $n$-th agent. Our final result $Q(o^n, u^n)$ can be obtained by taking \texttt{softmax} function for $Q^{\textsf{ally}}(o^{n},u^{n})$ and  $Q^{\textsf{enemy}}(o^{n},u^{n})$, and also adding weights.

\section{Performance Evaluation}\label{sec:4}

This section consists of simulation setup (refer to Sec.~\ref{sec:41}) and results (refer to Sec.~\ref{sec:42}), respectively.

\subsection{Simulation Setup}\label{sec:41}

Our simulation is performed in Windows operating systems and this MADRL is implemented with Python with open library Pytorch/Pysc2/SMAC.
Among various maps provided by SMAC, our simulation-based evaluation is performed using the scenarios with 3 marines (called \textsf{Simulation-3m}) and 8 marines (called \textsf{Simulation-8m}) where the level/difficulty of the game is set to $7$ (i.e., very hard). 
For exploit-and-exploration, $\epsilon$-greedy is used where $\epsilon=0.3$~\cite{levine2013guided}.
As illustrated in Fig.~\ref{fig:2}, actor and critic neural architectures are CSGA-policy and QMIX, respectively. The corresponding training is conducted with temporal difference~\cite{tesauro1995temporal} where the number of training iteration is $100,000$. 


\begin{figure*}[t!]
\centering
\setlength{\tabcolsep}{2pt}
\renewcommand{\arraystretch}{0.2}
\includegraphics[page=1, width=0.88\linewidth]{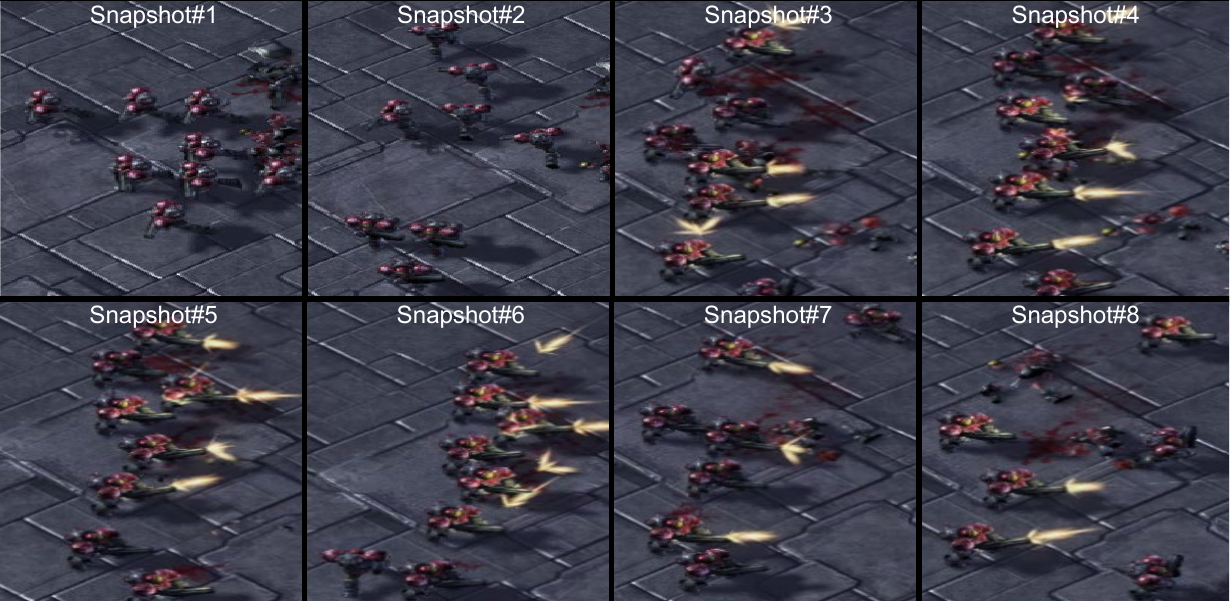}
\caption{MADRL performance evaluation results with \textsf{Simulation-8m} (8 marines vs. 8 marines).}
\label{fig:6}
\end{figure*}

\begin{figure*}[h!]
\centering
\setlength{\tabcolsep}{2pt}
\renewcommand{\arraystretch}{0.2}
\includegraphics[page=1, width=0.88\linewidth]{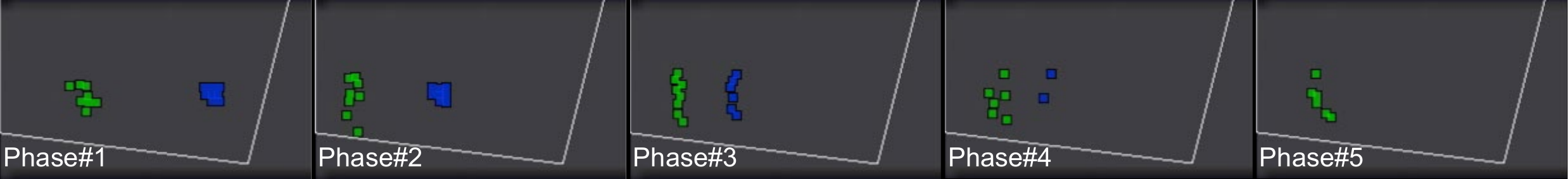}
\caption{The locations of agents in \textsf{Simulation-8m} (green: agents, blue: enemies).}
\label{fig:7}
\end{figure*}

\subsection{Simulation Results}\label{sec:42}

The SMAC simulation results in terms of \textsf{Simulation-3m} and \textsf{Simulation-8m} are presented in Fig.~\ref{fig:5}, Fig.~\ref{fig:6}, and Fig.~\ref{fig:7}. 
In the results, red units and blue units are for our ally agents and enemy agents (where the enemy is set to the level $7$ which means \textit{very hard}), respectively. 
The performance evaluation results for \textsf{Simulation-3m} and \textsf{Simulation-8m} are in following Sec.~\ref{sec:421} and Sec.~\ref{sec:422}, respectively. 
Furthermore, the demo video is available on~\cite{youtube-demo}.

\subsubsection{Simulation Results (3 marines)}\label{sec:421}
The six snapshots in Fig.~\ref{fig:5} are time-series representations of multi-agent actions for the case of \textsf{Simulation-3m}. 
In Fig.~\ref{fig:5}, Snapshot \#1 shows the situation before combat. Snapshots from \#2 to \#4 is for focused shooting and thus we can observed that one enemy is eliminated. In Snapshot\#5, one of our ally marines moves up because its own health condition is not good (thus, it wants to avoid the combat in order to keep its own health condition). In Snapshot \#6, the ally marine who moved up in Snapshot \#5 is back and attacks the enemies, and finally all enemies are eliminated. In our entire $20$ simulations, we confirm that our ally marines are all alive (i.e., winning rate: $100$\,\%).

\subsubsection{Simulation Results (8 marines)}\label{sec:422} 
The eight snapshots in Fig.~\ref{fig:6} are time-series representations of multi-agent actions for the case of \textsf{Simulation-8m}. 
In addition, Fig.~\ref{fig:7} shows the positions of ally units and enemy units. 
During the early stages (i.e., the Snapshots from \#1 to \#3 in Fig.~\ref{fig:6} and the Phases from \#1 to \#2 in Fig.~\ref{fig:7}), our ally marines move back, arrange combat formation, and lure the enemy units into our desired position.
In the middle stages (i.e., the Snapshots from \#4 to \#8 in Fig.~\ref{fig:6} and the Phases from \#3 to \#4 in Fig.~\ref{fig:7}), our ally units attack the enemy units, and the ally units who health conditions are not good move back. 
Lastly, we can observe that all enemy units are eliminated in the Phase \#5 in Fig.~\ref{fig:7}·


\section{Concluding Remarks}\label{sec:5}
This paper proposes a novel multi-agent deep reinforcement learning (MADRL) algorithm that controls multiple agents in real-time strategy (RTS) games. 
Among them, our proposed MADRL-based algorithm is fundamentally based on distributed computation and it is the combination of categorized state graph attention policy (CSGA-policy) and QMIX neural architectures. The CSGA-policy is the for state categorization and graph attention where the graph attention is used for learning the relationship among agents. The training is performed with QMIX which is centralized during training whereas it is distributed among agents during inference for individual action taking.
Our performance evaluation results verify that the proposed algorithm works well in \textit{StarCraft II} environment with various settings.

\section*{Acknowledgment}
This research was supported in part by Electronics and Telecommunications Research Institute (ETRI) grants 19YE1410 and Air Force Office of Scientific Research (AFOSR) grants FA2386-19-1-4020. S. Yi and J. Kim are the corresponding authors of this paper.

\bibliographystyle{IEEEtran}  
\bibliography{reference}

\end{document}